%% file: icons21.tex
\documentclass[sigconf]{acmart}
\AtBeginDocument{%
  \providecommand\BibTeX{{%
    \normalfont B\kern-0.5em{\scshape i\kern-0.25em b}\kern-0.8em\TeX}}}

\setcopyright{acmcopyright}
\copyrightyear{2021}
\acmYear{2021}
\acmDOI{10.1145/1122445.1122456}

\acmConference[ICONS '21]{ICONS '18: ACM International Conference on Neuromorphic Systems}{July 27--29, 2021}{Virtual Conference}
\acmBooktitle{ICONS '21: ACM International Conference on Neuromorphic Systems,
  June 03--05, 2018, Virtual Conference}
\acmPrice{15.00}
\acmISBN{978-1-4503-XXXX-X/18/06}

\usepackage[redeflists]{IEEEtrantools}
\newcommand{\sm}{\text{{SpiNeMap}}}{}
\newcommand{\pc}{\text{{PyCARL}}}{}
\newcommand{\esl}{\text{{DecomposedSNN}}}{}
\usepackage{amsmath}
\usepackage{multirow}
\usepackage{gensymb}
\usepackage{xcolor}
\usepackage[caption=false]{subfig}
\usepackage[linesnumbered,ruled]{algorithm2e}

\SetCommentSty{mycommfont}
\SetKwRepeat{Do}{do}{while}
\usepackage{threeparttable}
\newcommand{\mr}[1]{\textcolor{black}{#1}}
\usepackage{pifont}

\newcommand{\ineq}[1]{\footnotesize$#1$\normalsize}{}
\newtheorem{Tlemma}{Lemma}

\newtheorem{Tdef}{Definition}

\newcommand{\tech}{NeuroXplorer}{}

\begin{document}
\bstctlcite{IEEEexample:BSTcontrol}

\title{\tech{} 1.0: A Fast and Extensible Framework for \\Spiking Neural Network Simulation and Hardware Mapping}

\title{\tech{} 1.0: An Extensible Framework for Functional Simulation and Architectural Exploration with \\ Spiking Neural Network}

\title{\tech{} 1.0: An Extensible Framework for Architectural Exploration with Spiking Neural Networks}

\author{Adarsha Balaji}
\email{ab3586@drexel.edu}
\affiliation{%
  \institution{Drexel University}
  \city{Philadelphia}
  \state{PA}
  \country{USA}
  \postcode{19104}
}

\author{Shihao Song}
\email{ss3695@drexel.edu}
\affiliation{%
  \institution{Drexel University}
  \city{Philadelphia}
  \state{PA}
  \country{USA}
  \postcode{19104}
}

\author{Twisha Titirsha}
\email{tt624@drexel.edu}
\affiliation{%
  \institution{Drexel University}
  \city{Philadelphia}
  \state{PA}
  \country{USA}
  \postcode{19104}
}

\author{Anup Das}
\email{anup.das@drexel.edu}
\affiliation{%
  \institution{Drexel University}
  \city{Philadelphia}
  \state{PA}
  \country{USA}
  \postcode{19104}
}

\author{Jeffrey L. Krichmar}
\email{jkrichma@uci.edu}
\affiliation{%
  \institution{University of California, Irvine}
  \city{Irvine}
  \state{CA}
  \country{USA}
  \postcode{92697}
}

\author{Nikil Dutt}
\email{dutt@ics.uci.edu}
\affiliation{%
  \institution{University of California, Irvine}
  \city{Irvine}
  \state{CA}
  \country{USA}
  \postcode{92697}
}

\author{James Shackleford}
\email{jas64@drexel.edu}
\affiliation{%
  \institution{Drexel University}
  \city{Philadelphia}
  \state{PA}
  \country{USA}
  \postcode{19104}
}

\author{Nagarajan Kandasamy}
\email{nk78@drexel.edu}
\affiliation{%
  \institution{Drexel University}
  \city{Philadelphia}
  \state{PA}
  \country{USA}
  \postcode{19104}
}

\author{Francky Catthoor}
\email{Francky.Catthoor@imec.be}
\affiliation{%
  \institution{Imec}
  \city{Leuven}
  \state{}
  \country{Belgium}
  \postcode{3001}
}

\renewcommand{\shortauthors}{Balaji, et al.}

\begin{abstract}
    \input{sections/abstract}
\end{abstract}

\ccsdesc[500]{Hardware~Neural systems}
\ccsdesc[500]{Hardware~Emerging languages and compilers}
\ccsdesc[500]{Hardware~Emerging tools and meth\-odologies}

\keywords{Spiking Neural Networks (SNN), Neuromorphic Computing, Non Volatile Memory (NVM), Platform-Based Design, Hardware-Software Co-Design, Design-Technology Co-Optimization}


\maketitle

\section{Introduction}\label{sec:introduction}
\input{sections/introduction}

\section{\tech{}: High-Level Design}\label{sec:high_level_overview}

\input{sections/overview}

\section{Detailed Design of System Software}\label{sec:detailed_design}

\input{sections/detailed_design}

\section{Detailed Design of Neuromorphic Hardware Simulator}\label{sec:hardware_simulator}
\input{sections/simulator}

\section{Evaluation of \tech{}}\label{sec:evaluation}

\input{sections/evaluation}

\section{Conclusions}\label{sec:conclusions}
\input{sections/conclusions}

\bibliographystyle{IEEEtranSN}
\bibliography{commands,disco,external}

\end{document}

%% file: sections/abstract.tex
Recently, both industry and academia have proposed many different neuromorphic architectures to execute applications that are designed with Spiking Neural Network (SNN). 
Consequently, there is a growing need for an extensible 
simulation framework
that can perform architectural explorations with SNNs, including both platform-based design of today's hardware, and hardware-software co-design and design-technology co-optimization of the future. We present \tech{}, a fast and extensible framework that is based on a generalized template for modeling a neuromorphic architecture that can be infused with the specific details of a given hardware and/or technology. \tech{} can perform both low-level cycle-accurate architectural simulations and high-level analysis with data-flow abstractions. \tech{}'s optimization engine can incorporate hardware-oriented metrics such as energy, throughput, and latency, as well as SNN-oriented metrics such as inter-spike interval distortion and spike disorder, which directly impact SNN performance.
We demonstrate the architectural exploration capabilities of \tech{} through case studies with  many state-of-the-art machine learning models.

%% file: sections/introduction.tex
The term neuromorphic computing was coined in the 90s to describe integrated circuits that mimic the neuro-biological architecture of the central nervous system~\cite{mead1990neuromorphic}. 
These circuits employ variants of integrate-and-fire (I\&F) neurons~\cite{fusi1999collective} as computational units and analog weights as synaptic storage. 
I\&F neurons use spikes to encode information, where each spike is a voltage or current pulse in the physical world, typically of ms duration~\cite{maass1997networks}.
Recently, both industry and academia have proposed many different neuromorphic platforms to execute applications that are designed with Spiking Neural Network (SNN). Examples of such platforms include SpiNNaker~\cite{spinnaker}, Neurogrid~\cite{neurogrid}, TrueNorth~\cite{truenorth}, DYNAPs~\cite{dynapse}, Tianji~\cite{tianji}, Loihi~\cite{loihi}, and ODIN~\cite{odin}, among others~\cite{schuman2017survey}.

\begin{figure*}[t!]
	\centering
	\centerline{\includegraphics[width=2.01\columnwidth]{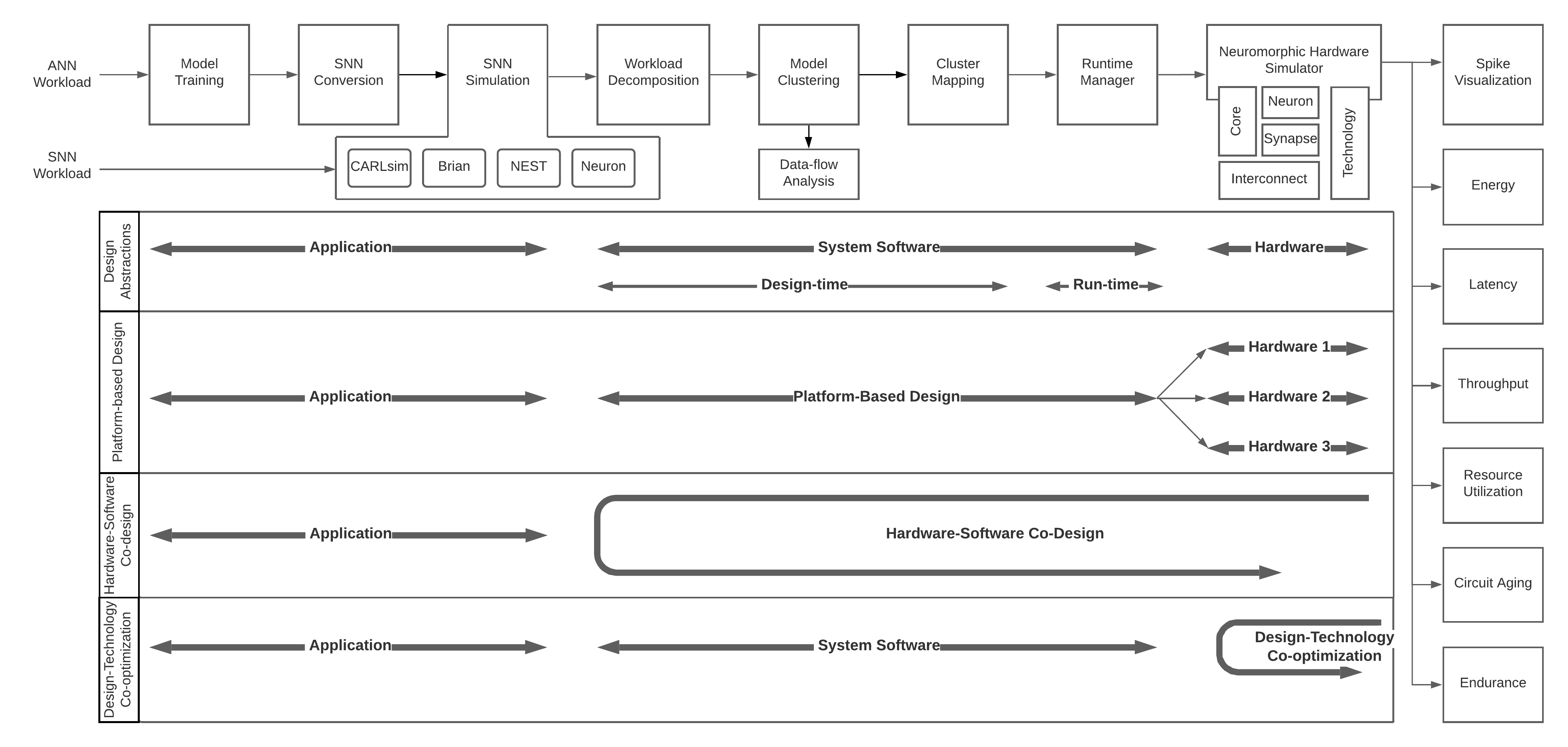}}
	\vspace{-10pt}
	\caption{High-level overview of \tech{}. The framework supports the following functionalities: 1) application quality exploration, 2) platform-based design, 3) hardware-software co-design, and 4) design-technology co-optimization.}
	\label{fig:overview}
\end{figure*}

To cope with the growing complexity of neuromorphic systems\footnote{The complexity of a neuromorphic system can be expressed in terms of the number of neurons and synapses, and their interconnection.}, challenges in integrating emerging non-volatile memory technologies, and faster time-to-market pressure, efficient design methodologies are needed~\cite{balaji2019design}. 
We highlight the following three key concepts
that are likely to address the design issues postulated above.
\begin{itemize}
    \item \textbf{Platform-based Design:} In this design methodology, a hardware platform is abstracted from its system software using the Application Programming Interface (API), making the hardware and software development orthogonal to allow more effective exploration of alternative solutions~\cite{keutzer2000system}. Platform-based design methodology facilitates the reuse of the system software for many different hardware platforms.
    \item \textbf{Hardware-Software Co-design:} In this design methodology, a hardware platform and its system software are concurrently designed to exploit their synergism in order to achieve system-level design objectives~\cite{de1997hardware}. The system software in this case is tailored for the hardware platform.
    \item \textbf{Design-Technology Co-optimization:} In this design meth\-odology, system-level design metrics are applied to explore the choices in hardware design and process technology to enable scaling at advanced technology nodes~\cite{yeric2013past}.
\end{itemize}

Consequently, there is a growing need for an extensible hardware simulator and an application mapper that can perform architectural explorations with SNNs, including platform-based design, hardware-software co-design, and design-technology co-optimi\-zation. 
We present \textbf{\tech{}}, a fast and extensible framework that is based on a \textit{generalized template} for modeling a neuromorphic architecture that can be infused with the specific details of a given hardware and/or technology.

\tech{} is released under the permissive MIT open license and it provides a user with the following high-level functionalities, which we will elaborate in the following sections.
\begin{itemize}
    \item A design optimization engine that can incorporate hardware design metrics such as energy, latency, throughput, and reliability, as well as SNN-oriented metrics such as inter-spike interval distortion and spike disorder.
    \item A generalized and optimized system software framework, facilitating mapping of SNN-based applications to different neuromorphic hardware platforms.
    \item A cycle-accurate model of neuromorphic hardware utilizing a generalized template, which can be configured with hardware- and technology-specific details from industrial and academic manufacturers of neuromorphic systems.
    \item A design space exploration framework using data flow abstractions to represent machine learning models for execution on neuromorphic hardware, allowing estimation of key system-level performance metrics early in the system design stage.
    \item A framework to analyze different technological alternatives for neuron and synapse circuits, and the impact of scaling in neuromorphic hardware, facilitating optimization of key system-level design metrics.
\end{itemize}

In addition to these architecture-centric functionalities, \tech{} also facilitates functional simulations via SNN simulators such as CARLsim~\cite{carlsim}, Brian~\cite{brian}, NEST~\cite{nest}, and Neuron~\cite{neuron}, supporting different degrees of neuro-biological details and learning modalities. Thus, \tech{} allows to explore the design-space of application performance alongside architecture development.

\tech{} is developed over a period of five years and is supported by three National Science Foundation research grants and one Department of Energy grant from the United States, and one Horizon 2020 research grant from Europe.

%% file: sections/overview.tex
Figure~\ref{fig:overview} illustrates the key components of \tech{}. At a high-level, \tech{} supports three layers of abstraction -- the \textbf{application} layer, the \textbf{system software} layer, and the \textbf{hardware} layer, similar to the abstractions in a classical von-Neumann computing system. Internally, the system software layer is divided into a \textbf{design-time} or \textbf{compile-time} methodology, where a machine learning model is converted into an intermediate form for mapping to a specific neuromorphic hardware, and a \textbf{run-time} methodology, which allocates hardware resources to admit and execute the model on the hardware. 
\tech{} can work with both {Artificial Neural Networks} (ANNs) and biology-inspired Spiking Neural Networks (SNNs).
\tech{} interfaces with ANN workloads that are specified in high-level frameworks such as Keras with TensorFlow backend~\cite{keras,tensorflow} and PyTorch~\cite{pytorch}. To map an ANN workload to an event-driven neuromorphic hardware, the workload is first converted to an SNN using the \textbf{SNN Conversion} unit and then, the SNN is simulated using the \textbf{SNN Simulation} unit of \tech{}.

SNN workloads can be specified in PyNN~\cite{pynn}, which is a Python interface to many SNN simulators such as CARLsim~\cite{carlsim}, Brian~\cite{brian}, NEST~\cite{nest}, and Neuron~\cite{neuron}. These simulators model neural functions at various levels of detail and therefore have different requirements for computational resources. User can also specify an SNN model directly using these simulators. \tech{} allows exploration of application quality using these simulators.

\tech{} incorporates the spike timing information obtained from simulating an SNN model with representative training data. Such information is used to map the model to the neuromorphic hardware using the system-software framework, which consists of \textbf{Workload Decomposition}, \textbf{Model Clustering}, \textbf{Cluster Mapping}, and \textbf{Runtime Management} units.

Without loss of generality, we describe \tech{} where ANN workloads are specified using Keras and SNN workloads using a combination of PyNN and PyCARL~\cite{pycarl}, a Python wrapper for SNN simulations using CARLsim. Additionally, we use our previously proposed SNN converter~\cite{jolpe18} for SNN conversion of ANN workloads in order to map them to hardware. \tech{} can be trivially \textbf{extended} to work with other SNN simulation tools such as GeNN~\cite{genn} and Spyketorch~\cite{spyketorch}, and with other SNN conversion approaches such as~\cite{midya2019artificial,rueckauer2016theory,rueckauer2018conversion}.

Figure~\ref{fig:overview} illustrates 
the three design methodologies supported by \tech{} -- 1) platform-based design, 2) hardware-software co-design, and 3) design-technology co-optimization.
We have used \tech{} to optimize for system-level design metrics, including energy~\cite{psopart,spinemap,twisha_energy}, latency~\cite{das2018dataflow,balaji2019ISVLSIframework}, throughput~\cite{dfsynthesizer}, resource utilization~\cite{esl20,adarsha_igsc}, circuit aging~\cite{balaji2019framework,reneu,song2020case,vts_das}, and endurance~\cite{twisha_endurance,twisha_thermal,espine}.


%% file: sections/detailed_design.tex
We now 
detail the system software of \tech{}. 

\subsection{Platform Description}
\begin{figure}[h!]%
    \centering
    \vspace{-20pt}
    \subfloat[A tile-based neuromorphic hardware~\cite{catthoor2018very}. A tile communicates spikes via the network switches (S) using a shared interconnect such as Networks-on-chip (NoC)~\cite{liu2018neu} and Segmented Bus~\cite{balaji2019exploration}. \label{fig:architecture}]{{\includegraphics[width=4.0cm]{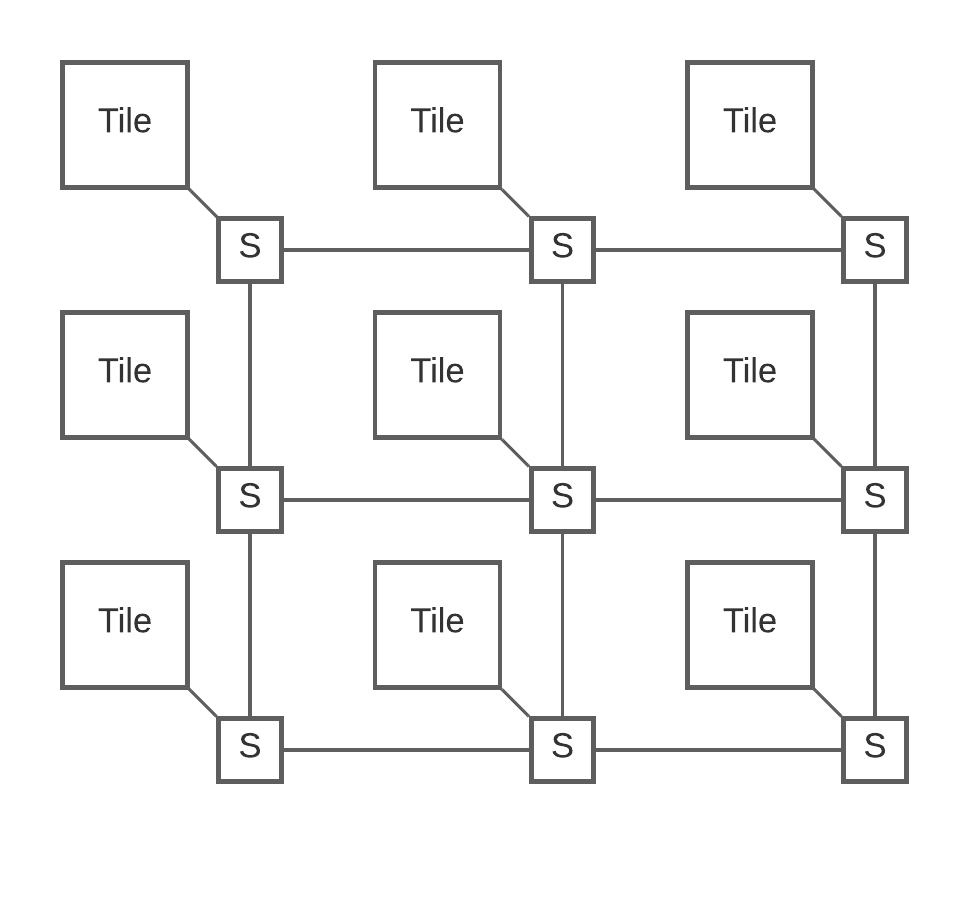} }}%
    \quad
    \subfloat[A crossbar architecture.
    Spikes are encoded into Address Event Representation (AER). A crossbar peripheral circuitry consists of AER encoder and decoder. \label{fig:crossbar}]{{\includegraphics[width=4.0cm]{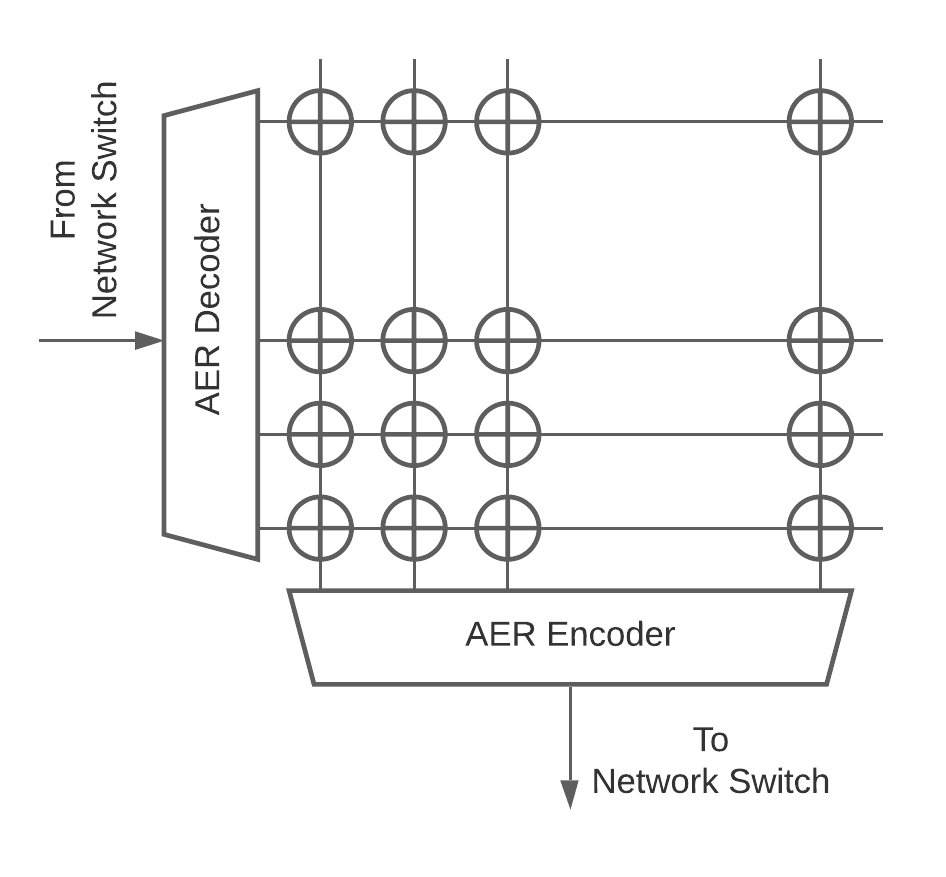} }}%
    \vspace{-10pt}
    \caption{A running example of a tile-based neuromorphic hardware. Each tile contains a neuromorphic core, which in its simplest form can be a crossbar.}%
    \label{fig:neuromorphic_architecture}%
\end{figure}

We 
consider
a tile-based neuromorphic hardware as shown in Figure~\ref{fig:architecture}. Each tile consists of a neuromorphic core, which can accommodate a certain number of neurons and synapses. 
A common
approach to implementing a neuromorphic core is one where the synaptic cells are organized in a two-dimensional grid, known as crossbar.
We illustrate a crossbar in Figure~\ref{fig:crossbar}.\footnote{Although \tech{} provides a generalized template for crossbar-based neuromorphic tiles, \tech{} can be easily extended to support many different types of processing elements such as~\cite{qiao2015reconfigurable,ma2017darwin}.} 

Typically, system designers limit the size of a crossbar to reduce energy consumption \footnote{Energy consumption in a crossbar scales proportional to $M^2$, where $M$ is the input/output dimension of a crossbar.} and mitigate the high parasitic voltage drops within a crossbar (see Figure~\ref{fig:parasitics}). Therefore, a large machine learning model must be partitioned into \textit{local synapses}, those that map within the crossbar of a tile, and \textit{global synapses}, those that map on the shared interconnect~\cite{psopart}. To effectively address this partitioning, \tech{}'s system software performs the following four key functionalities to map a machine learning workload to the hardware: workload decomposition, model clustering, cluster mapping, and runtime.
We now describe these functions.

\subsection{Workload Decomposition}\label{sec:decomposition}
\begin{figure}[h!]
	\centering
	\vspace{-10pt}
	\centerline{\includegraphics[width=0.99\columnwidth]{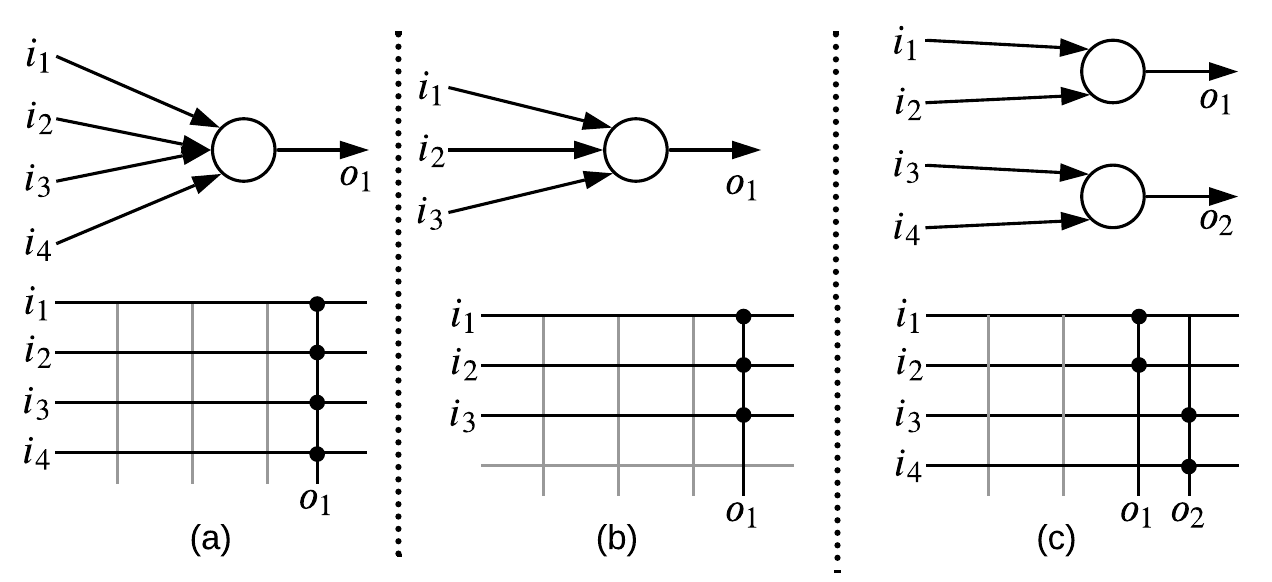}}
	\vspace{-10pt}
	\caption{Example mapping of a) one 4-input, b) one 3-input, and c) two 2-input neurons to a $4 \times 4$ crossbar.}
	\label{fig:crossbar_mapping}
\end{figure}

We note that each $N \times N$ crossbar in a tile can accommodate up to \ineq{N} pre-synaptic connections per post-synaptic neuron, with typical value of \ineq{N} set between 128 (in DYNAPs) and 256 (in TrueNorth).
Figure~\ref{fig:crossbar_mapping} illustrates an example of mapping a) one 4-input, b) one 3-input, and c) two 2-input neurons on a $4 \times 4$ crossbar. Unfortunately, neurons with more than 4 pre-synaptic connections per post-synaptic neuron cannot be mapped to this crossbar.

We take the example architecture of DYNAPs, where each crossbar can accommodate a maximum of 128 pre-synaptic connections.
In many complex machine learning models such as LeNet, AlexNet, VGG, ResNet, and DenseNet, the number of pre-synaptic connections per post-synaptic neuron is much higher than what a crossbar in DYNAPs can accommodate. 

\begin{figure}[h!]
	\centering
	\centerline{\includegraphics[width=0.99\columnwidth]{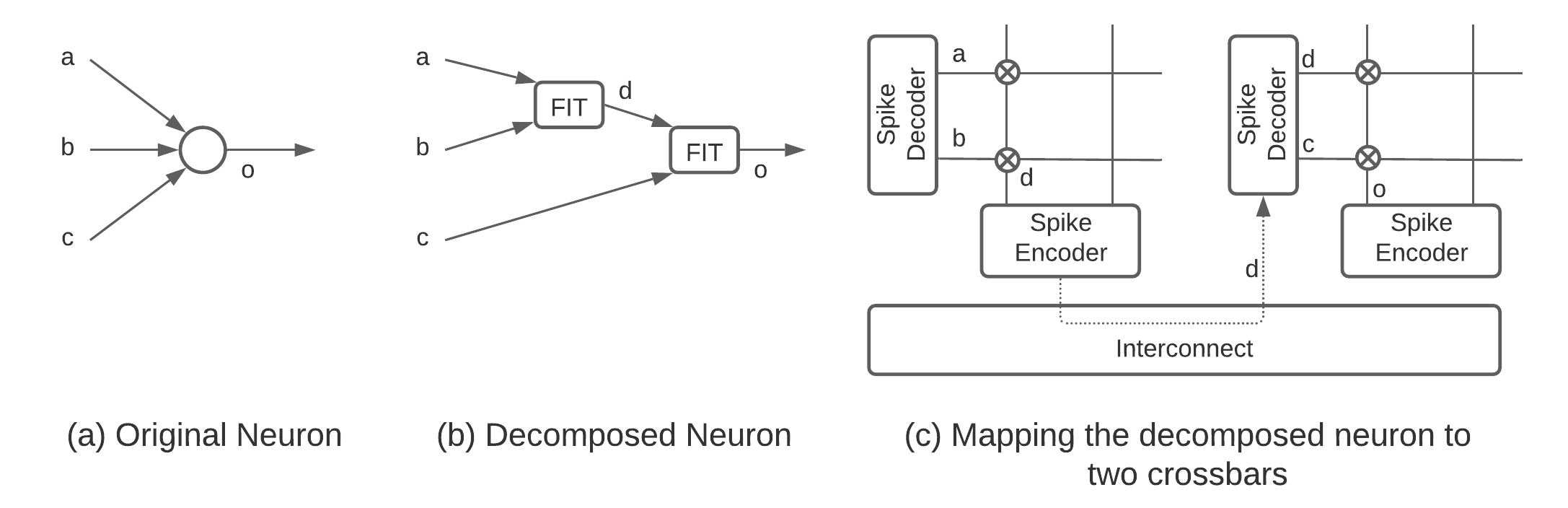}}
	\vspace{-10pt}
	\caption{\mr{Illustrating the decomposition of a 3-input neuron (a) to a sequence of FIT neural units (b). The mapping of the FIT units to two 2x2 crossbars is shown in (c).}}
	\vspace{-10pt}
	\label{fig:decomposition_demo}
\end{figure}

To address this resource limitation, we have previously proposed workload decomposition, which exploits the firing principle of LIF neurons, decomposing each neuron with many pre-synaptic connections into a sequence of homogeneous fanin-of-two (FIT) neural units~\cite{esl20}. Figure~\ref{fig:decomposition_demo} illustrates the spatial decomposition using a small example of a 3-input neuron shown in Figure~\ref{fig:decomposition_demo}(a). We consider the mapping of this neuron to 2x2 crossbars. Since each crossbar can accommodate a maximum of two pre-synaptic connections per neuron, the example 3-input neuron cannot be mapped to the crossbar directly. The most common solution is to eliminate a synaptic connection, which may lead to accuracy loss. Figure~\ref{fig:decomposition_demo}(b) illustrates the decomposition mechanism, where the 3-input neuron is implemented using two FIT neural units connected in sequence as shown in Figure~\ref{fig:decomposition_demo}(b). Each FIT unit is similar to a 2-input neuron and it exploits the leaky integrate behavior in hardware to maintain the functional equivalence between Figures~\ref{fig:decomposition_demo}(a) and \ref{fig:decomposition_demo}(b). Finally, Figure~\ref{fig:decomposition_demo}(c) illustrates the mapping of the decomposed neuron utilizing two 2x2 crossbars. The functionality of the FIT neural units is implemented using the synaptic cells of the two crossbars.

Workload decomposition is an optional function in \tech{}. If this function is disabled, a machine learning model is directly fed to the clustering step. In this case, some of the pre-synaptic connections may need to be eliminated to fit onto a crossbar, which could potentially lead to accuracy loss.

\subsection{Model Clustering}
 In the model clustering step, a large and complex machine learning model is partitioned into clusters, where each cluster consists of a fraction of the neurons and synapses of the original model that can fit onto the resources of a neuromorphic core.

\begin{figure}[h!]
	\centering
	\centerline{\includegraphics[width=0.5\columnwidth]{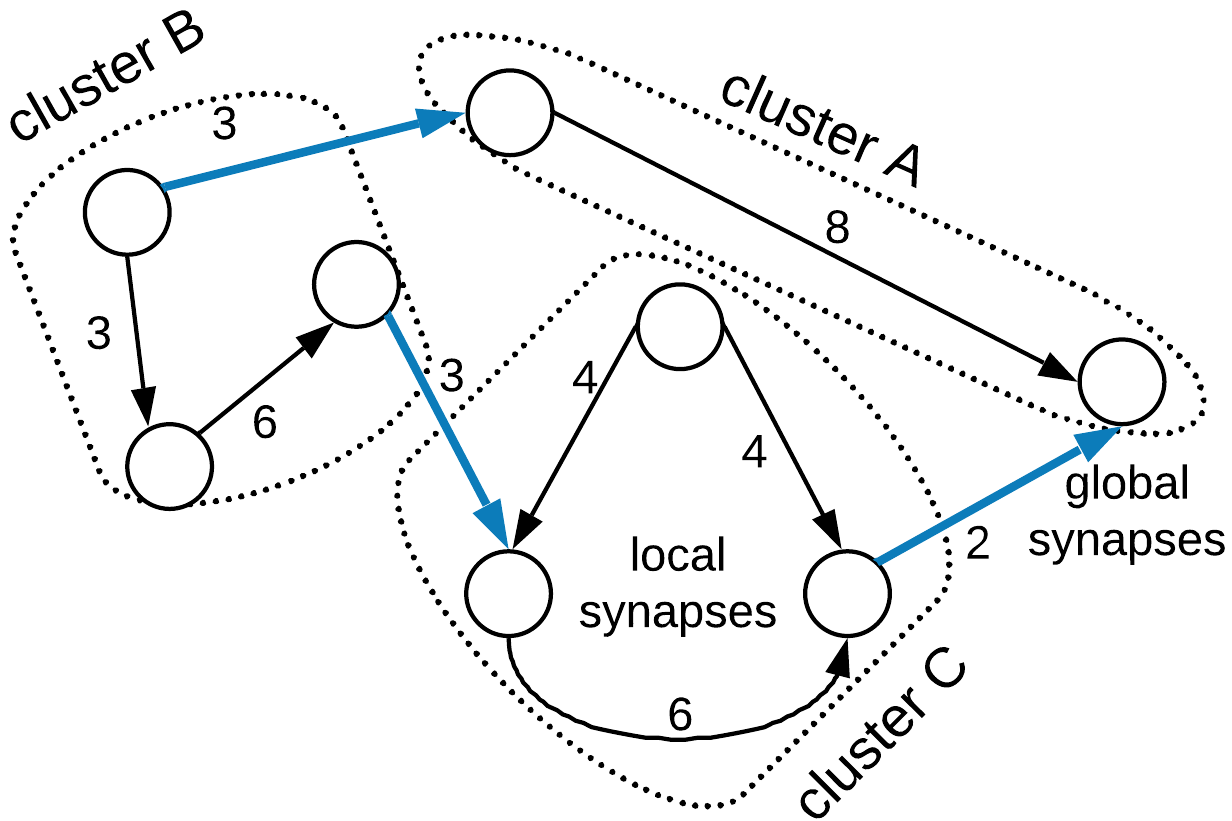}}
	\caption{An SNN partitioned into three clusters.}
	\label{fig:cluster_demo}
\end{figure}

 Figure \ref{fig:cluster_demo} illustrates an SNN partitioned into three clusters A, B, and C. The number of spikes communicated between a pair of neurons is indicated on its synapse. We indicate the local synapses (those that are within each cluster) in black and the global ones (those that are between clusters) in blue in this figure.

The SNN partitioning problem is essentially a graph partitioning problem, which is NP-complete.
Therefore, heuristics are typically used to find solutions. \tech{} currently supports two heuristics -- Particle Swarm Optimization (PSO)~\cite{pso} and Kernighan-Lin Graph Partitioning algorithm~\cite{kernighan1970efficient}. \tech{} uses these heuristics to minimize 1) the number of clusters (as in~\cite{esl20}), which reduces the hardware requirement, and 2) the number of inter-cluster spikes (as in \cite{psopart,spinemap}), which reduces both energy and latency when the machine learning model is mapped to hardware.
\tech{} can be easily extended to use other heuristics such as Hill Climbing~\cite{hill_climbing} and Simulated Annealing~\cite{simulated_annealing}, as well as other optimization objectives such as application quality and hardware reliability.

\begin{figure}[h!]%
    \centering
    \subfloat[VGG Convolution Neural Network (CNN).\label{fig:vgg}]{{\includegraphics[width=8.7cm]{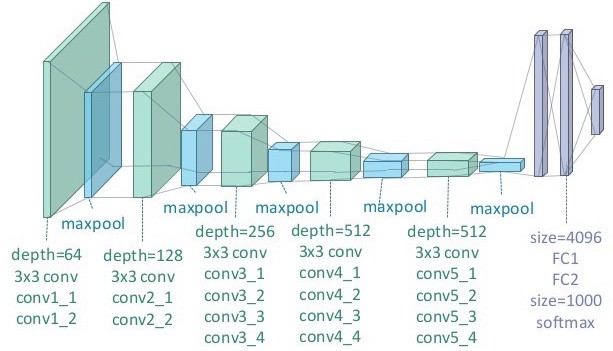} }}%
    \quad
    \subfloat[First 10 clusters of VGG generated using~\cite{spinemap}.\label{fig:cluster_vgg}]{{\includegraphics[width=6.7cm]{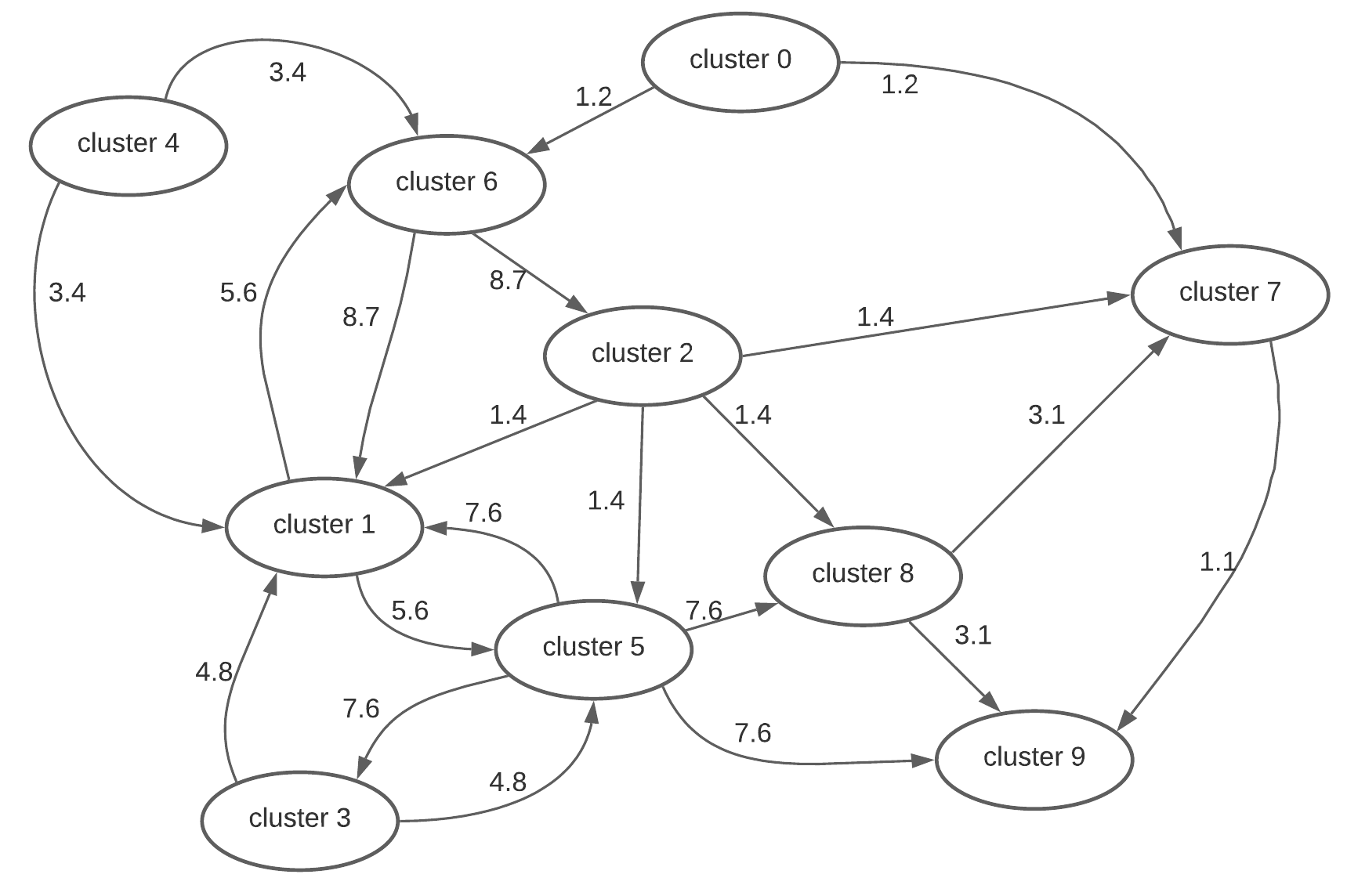} }}%
    \caption{Trained VGG model and its clusters generated using model partitioning tool such as SpiNeMap~\cite{spinemap}.}%
    \label{fig:vggnet_example}%
\end{figure}

Figure~\ref{fig:vgg} shows the architecture of VGG for CIFAR-10 classification. Figure~\ref{fig:cluster_vgg} shows the first 10 clusters generated using \sm{}~\cite{spinemap}.\footnote{SpiNeMap~\cite{spinemap} generates 95,452 clusters from the VGG model trained on CIFAR-10 dataset. For simplicity, we illustrate only the first 10 clusters and their interconnection.} The figure illustrates the connections between these clusters, with the number on edge representing the average number of spikes communicated between the source and destination clusters when processing an image during inference. The inter-cluster links are the global synapses for mapping purposes.

\subsection{Cluster Mapping}
\begin{figure}[h!]%
    \centering
    \vspace{-10pt}
    \subfloat[An example clustered SNN with three clusters -- A, B, and C.\label{fig:snn_ex}]{{\includegraphics[width=3cm]{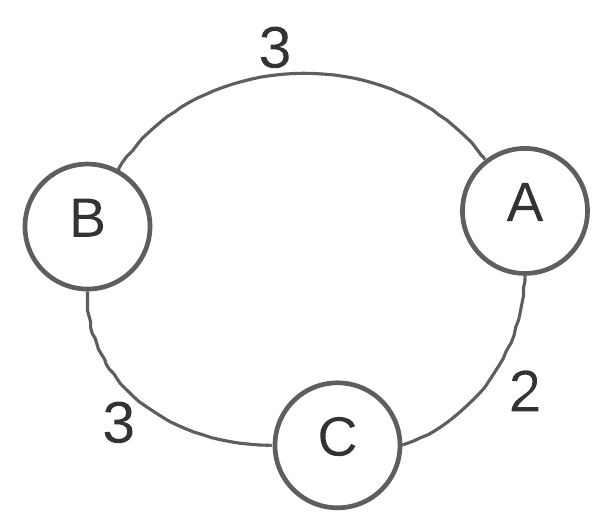} }}%
    \qquad
    \subfloat[An example placement of the three clusters to a mesh architecture. The three different colors (orange, blue, and green) indicate the the distance (in hops) spikes travel in the mesh architecture between source and destination clusters.
    \label{fig:placement_ex}]{{\includegraphics[width=4cm]{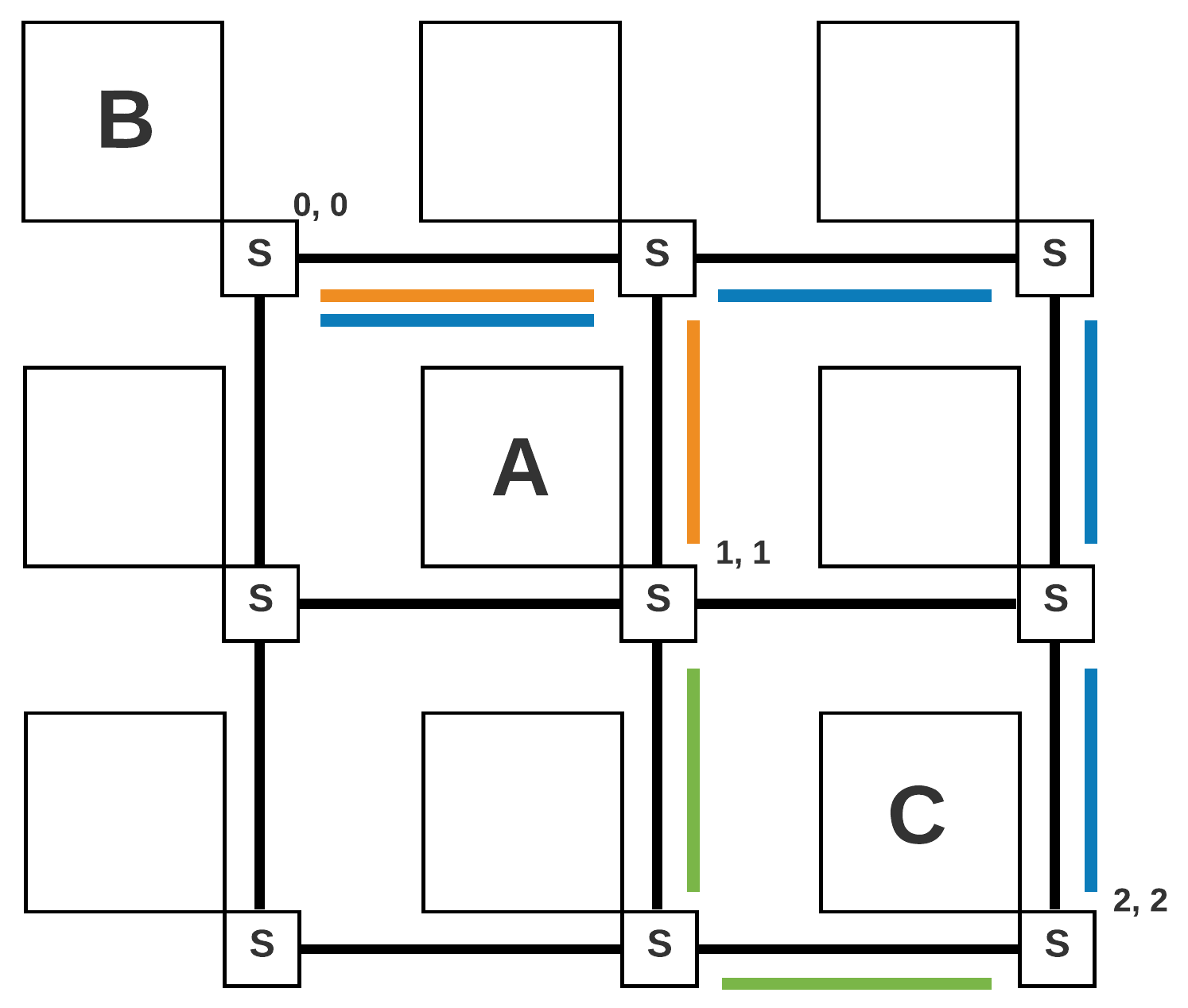} }}%
    \caption{Example of mapping a clustered SNN on a mesh architecture.}%
    \label{fig:comm_energy_compute_ex}%
\end{figure}

The cluster mapping step of \tech{} is used to reserve computing resources of the hardware for a given machine learning model and execute the model by placing its clusters onto the physical cores.
Figure~\ref{fig:placement_ex} illustrates the placement of a clustered SNN of Figure~\ref{fig:snn_ex} to a neuromorphic hardware with 9 cores organized in a mesh architecture. The position of each core in the hardware is specified by a pair of Cartesian coordinates. In this example, cluster A is placed at coordinate (1,1), cluster B at (0,0), and cluster C at (2,2). All spikes between A and B, and between A and C are communicated via two interconnect segments and one hop, while all spikes between B and C are communicated via four interconnect segments and three hops. Clearly, the latency and energy on the shared interconnect depends on the placement of the clusters on the physical cores located in the Cartesian coordinate system.

\tech{} uses an instance of PSO to optimize the placement of clusters of a machine learning model to the physical cores of the hardware, improving both latency and energy consumption. The placement solution of \tech{} aims to place the clusters that communicate the most to nearby cores. \tech{} can be extended to use other placement heuristics.

\subsection{Runtime Manager}
\begin{figure}[h!]
 	\centering
    \vspace{-6pt}
 	\centerline{\includegraphics[width=0.99\columnwidth]{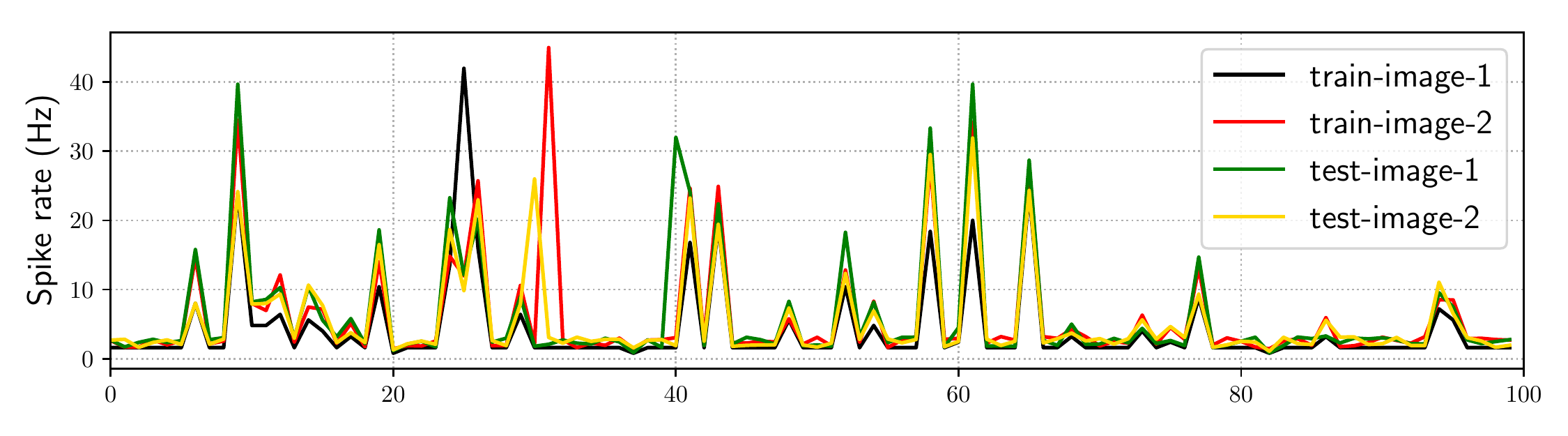}}
 	\vspace{-10pt}
 	\caption{Spike rate of 100 randomly-selected neurons in AlexNet for 2 training images and 2 test images.}
    \vspace{-10pt}
 	\label{fig:alexnet}
\end{figure}

To illustrate the significance of a run-time manager, Figure~\ref{fig:alexnet} plots the spike firing rate of 100 randomly-selected neurons in AlexNet~\cite{alexnet}, a state-of-the-art CNN used for Imagenet classification. We report results for two randomly-selected training and test images.

We observe that spike firing rates of neurons depend on the image presented to the AlexNet CNN. Therefore, energy and reliability improvement strategies based on design-time analysis with training examples may not be optimal when they are applied at run-time to process in-field data.
Therefore, in addition to cluster placement when admitting a machine learning model to hardware, \tech{} also supports monitoring key performance statistics collected from the hardware during model execution. 
Such statistics can uncover bottlenecks, allowing improving system-level metrics such as energy~\cite{balaji2020run} and circuit aging~\cite{balaji2019framework,ncrtm} through remapping of the neurons and synapses to the hardware.

%% file: sections/simulator.tex
\begin{figure}[h!]
	\centering
	\centerline{\includegraphics[width=0.99\columnwidth]{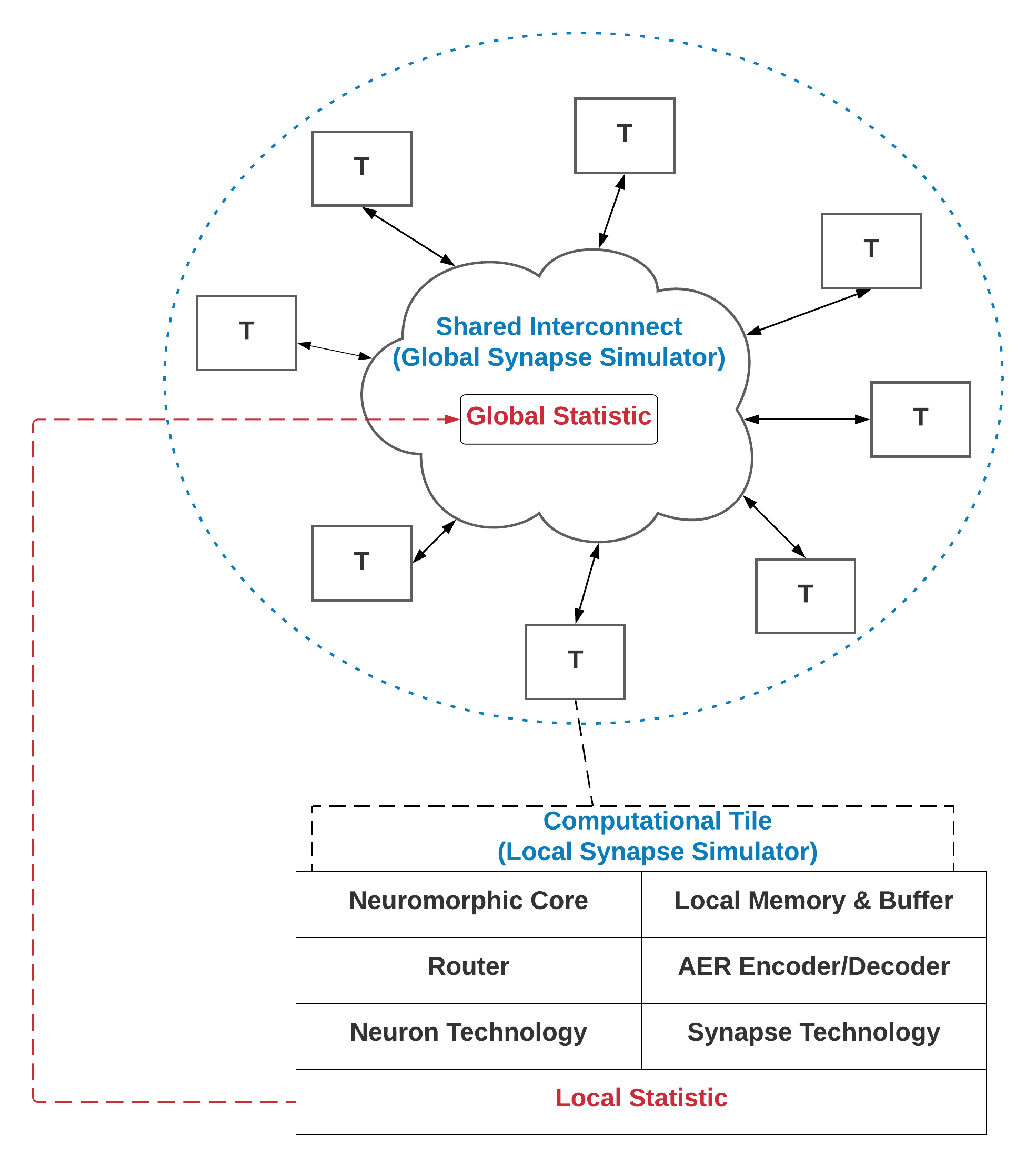}}
	\vspace{-10pt}
	\caption{Architecture simulator of \tech{}.}
	\label{fig:simulator_overview}
\end{figure}

Figure~\ref{fig:simulator_overview} shows the high-level overview of the proposed neuromorphic hardware simulator, which facilitates cycle-accurate simulation of the interconnect and the processing tiles. Each tile models 1) a processing element, which is a neuromorphic core, 2) a router for routing spike AER packets on the shared interconnect, 3) a local memory to store cluster parameters, 4) buffer space for spike packets, and 5) AER encoder and decoder. \tech{}'s hardware simulator can perform exploration with transistor technologies such as CMOS and FinFET that are used for the neurons and the peripheral circuitry in each tile, and Non-Volatile Memory (NVM) technologies such as Phase-Change Memory (PCM)~\cite{Burr2017}, Oxide-Based Resistive Random Access Memory (OxRRAM)~\cite{mallik2017design}, Ferroelectric RAM (FeRAM)~\cite{mulaosmanovic2017novel}, and Spin-Transfer Torque Magnetic or Spin-Orbit-Torque RAM (STT/ SoT-MRAM)~\cite{vincent2015spin} used for synaptic weight storage.\footnote{Beside neuromorphic computing, NVMs are also used as main memory in conventional computers to improve performance and energy efficiency~\cite{palp,mneme,datacon,hebe,shihao_igsc}.} We now describe the simulator. 

\subsection{Cycle-Accurate Interconnect Simulator}
\begin{figure}[h!]
	\centering
	\vspace{-10pt}
	\centerline{\includegraphics[width=0.55\columnwidth, angle =270]{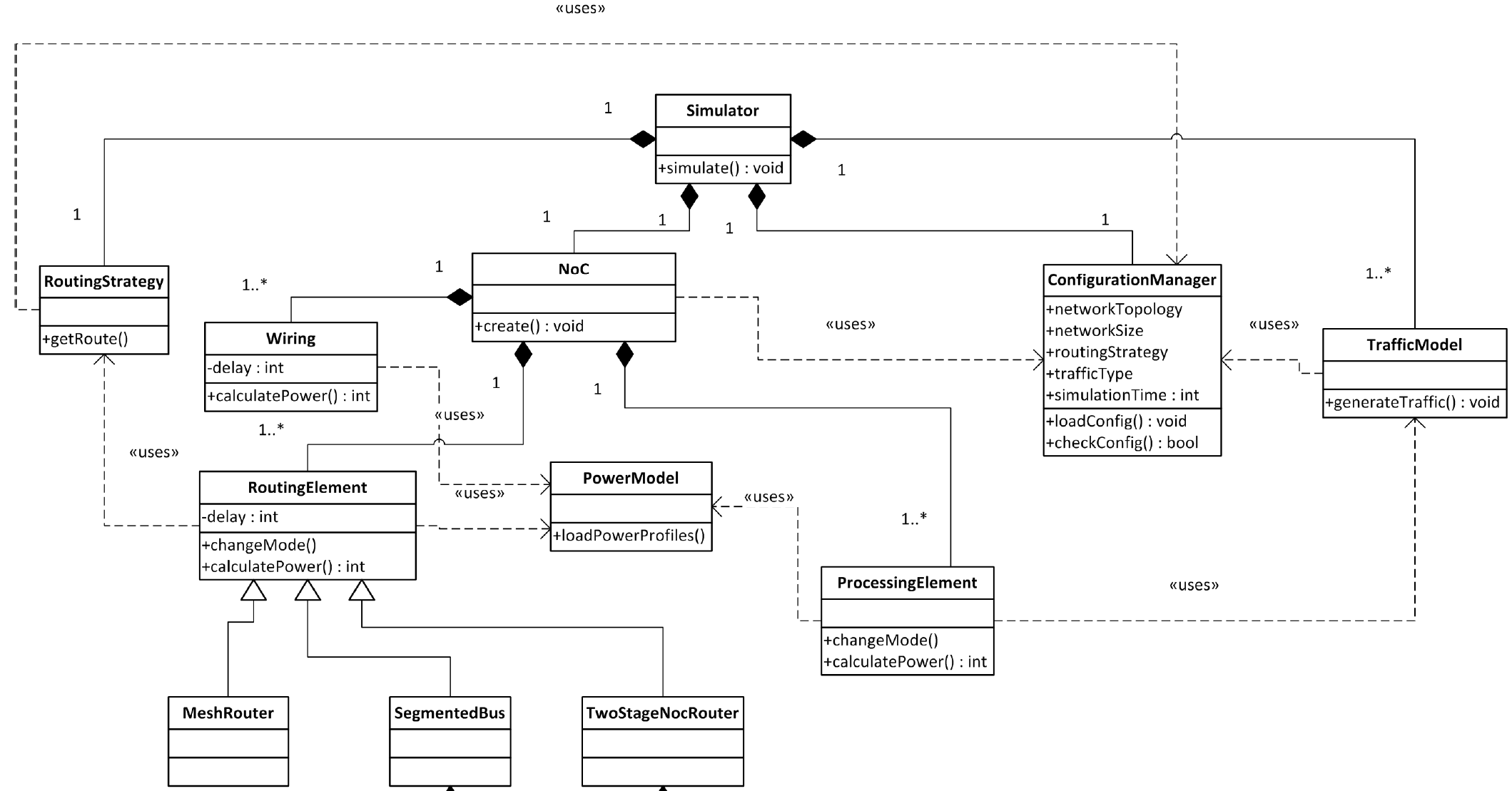}}
	\vspace{-10pt}
	\caption{Class diagram of \tech{}'s hardware simulator using UML convention.}
	\vspace{-10pt}
	\label{fig:class_diagram}
\end{figure}

Figure~\ref{fig:class_diagram} illustrates the internal architecture of the interconnect (global synapse) simulator of \tech{} in UML convention. Key components of this simulator are the following 
\begin{itemize}
    \item Spike Routing Strategy: This is the generalization class of the following routing strategies: Dyad, Negative First, North Last, Odd Even, Table-based, West-First, and XY.
    \item Spike Traffic Model: This is the generalization class of the following traffic models: Random, Transpose Matrix, Bit-Reversal, Butterfly, Shuffle, and Table-based.
    \item Configuration Manager: This is the generalization class for loading simulator parameters such as network topology, network size, traffic type, routing strategy, and simulation time.
\end{itemize}

\begin{figure}[h!]
	\centering
	\centerline{\includegraphics[width=0.8\columnwidth]{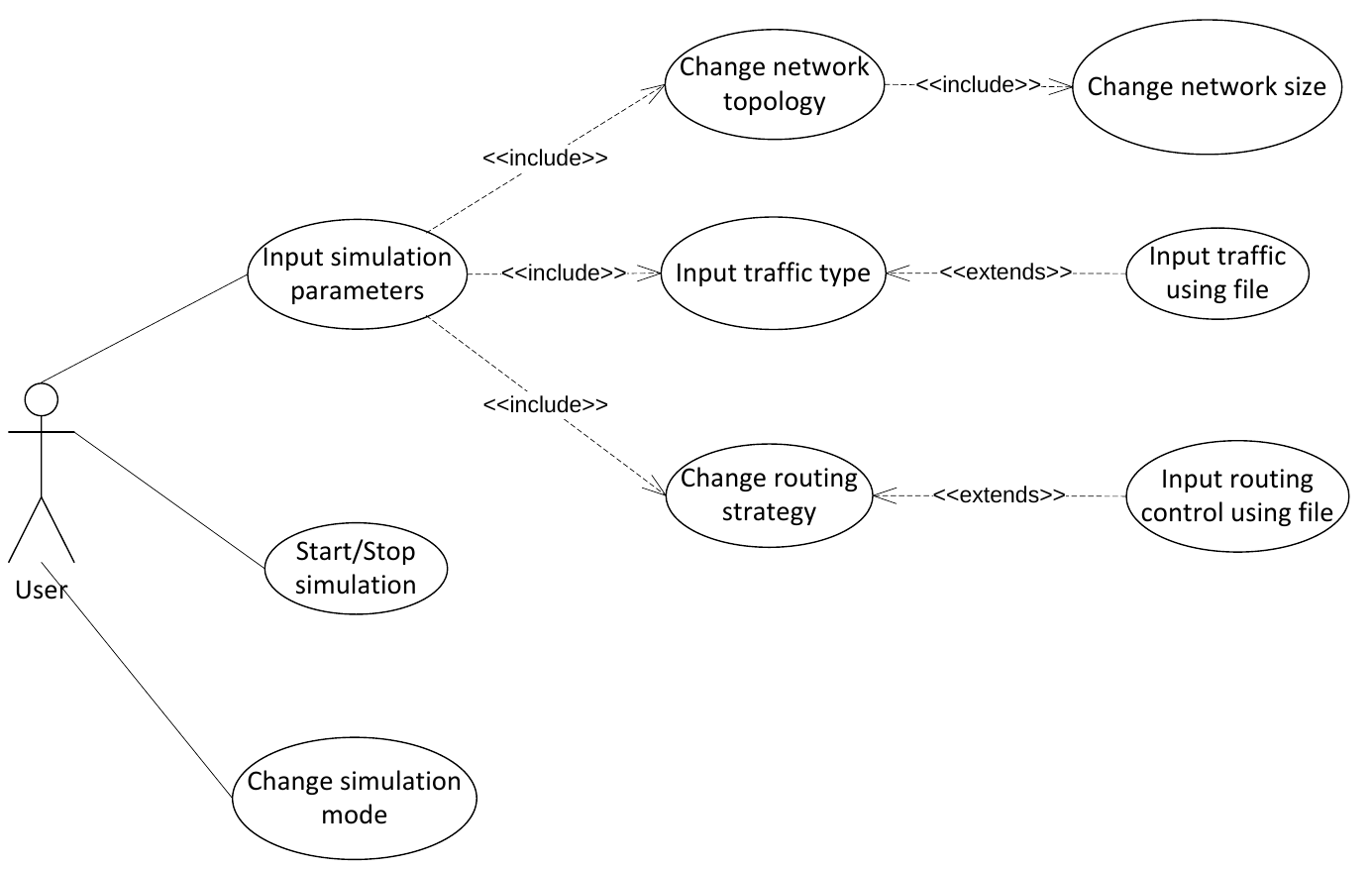}}
	\vspace{-5pt}
	\caption{Use-case diagram of the hardware simulator of \tech{}.}
	\label{fig:use_case_diagram}
\end{figure}

Figure~\ref{fig:use_case_diagram} shows the capabilities of the hardware simulator of \tech{}.
For example, when changing the network topology, the user can select between the three interconnect types: Mesh, Segmented bus, and Two-stage NoC. The user can also input spike traffic generated from the application-level simulator at the frontend of \tech{} to run hardware network simulation.

\subsection{Cycle-Accurate Tile Simulator}
\begin{figure}[h!]
	\centering
	\vspace{-5pt}
	\centerline{\includegraphics[width=0.8\columnwidth]{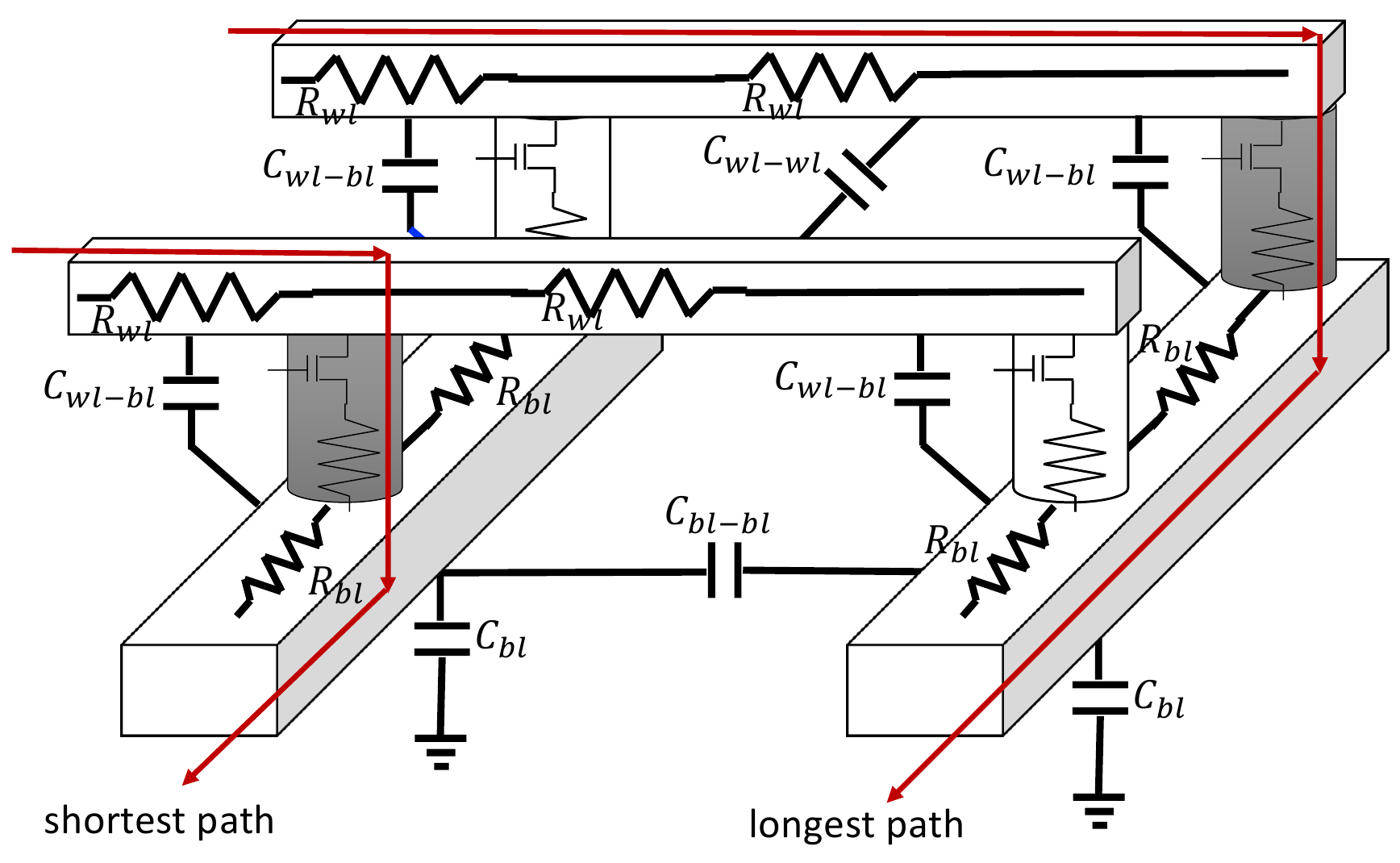}}
	\vspace{-10pt}
	\caption{Detailed model of a crossbar in \tech{}.}
	\vspace{-10pt}
	\label{fig:parasitics}
\end{figure}

On the computing tile front, \tech{} supports detailed model of crossbars with PCM and OxRRAM-based synaptic cells. 
Figure~\ref{fig:parasitics} shows the detailed circuit model of a crossbar in \tech{} with all of its parasitic components. Such parasitic components cause variable delays on the current paths inside the crossbar. 
For simplicity, we have only shown the current on the shortest and the longest paths in the crossbar, where the length of a current path is measured in terms of the number of parasitic elements on the path. Therefore, spike propagation delay through synapses on longer paths is higher than on shorter paths. \tech{} allows estimating these delays for a given process technology node.

Table~\ref{tab:parasitics} shows the template for specifying the parasitic components in \tech{} for a specific technology node.

\begin{table}[h!]
    \caption{Generalized template for specifying the parasitic components of a crossbar in \tech{}.}
	\label{tab:parasitics}
	\vspace{-10pt}
	\centering
	{\fontsize{6}{10}\selectfont
		\begin{tabular}{lp{5cm}}
			\hline
			$R_{wl}$ & unit wordline resistance\\
			$R_{bl}$ & unit bitline resistance\\
			\hline
			$C_{wl}$ & unit wordline capacitance\\
			$C_{bl}$ & unit bitline capacitance\\
			\hline
			$C_{wl-wl}$ & unit wordline to wordline capacitance\\
			$C_{wl-bl}$ & unit wordline to bitline capacitance\\
			$C_{bl-bl}$ & unit bitline to bitline capacitance\\
			\hline
	\end{tabular}}
	\vspace{-10pt}
\end{table}

\begin{figure}[h!]
	\begin{center}
		\vspace{-10pt}
		\includegraphics[width=0.69\columnwidth]{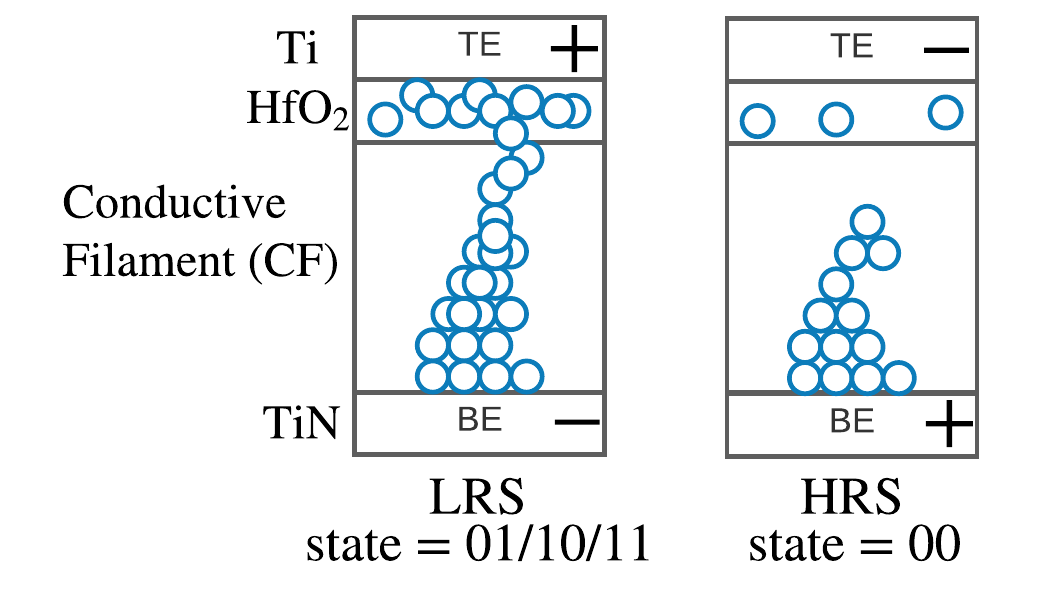}
		\vspace{-10pt}
		\caption{Operation of an RRAM cell with the $\text{HfO}_2$ layer sandwiched between the metals Ti (top electrode) and TiN (bottom electrode). The left subfigure shows the formation of LRS states with the formation of conducting filament (CF). This represents logic states 01, 10, and 11. The right subfigure shows the depletion of CF on application of a negative voltage on the TE. This represents the HRS state or logic 00.}
		\vspace{-10pt}
		\label{fig:RRAM}
	\end{center}
\end{figure}

On the technology front, we briefly discuss the OxRRAM technology, 
as an instance of a technology 
that can be used for the synaptic cell. 
An RRAM cell is composed of an insulating film sandwiched between conducting electrodes forming a metal-insulator-metal (MIM) structure (see Figure~\ref{fig:RRAM}). Recently, filament-based metal-oxide RRAM implemented with transition-metal-oxides such as HfO${}_2$, ZrO${}_2$, and TiO${}_2$ has received considerable attention due to their low-power and CMOS-compatible scaling.
Synaptic weights are represented as conductance of the insulating layer within each RRAM cell. To program an RRAM cell, elevated voltages are applied at the top and bottom electrodes, which re-arranges the atomic structure of the insulating layer. Figure~\ref{fig:RRAM} shows the High-Resistance State (HRS) and the Low-Resistance State (LRS) of an RRAM cell. 
In \tech{}, the RRAM cell can also be programmed into intermediate low-resistance states, allowing its multilevel operations. 
For instance, 
to implement two bits per synapse we can program the RRAM cell for one HRS and three LRS states.

\tech{} also supports implementing many variants of Integrate \& Fire (I\&F) neuron. 
Table~\ref{tab:hw_parameters} provides the template for specifying the parameters for a neuron and synaptic cell in a crossbar.
\begin{table}[h!]
    \caption{Generalized template for specifying the parameters of a neuron and synaptic cell in \tech{}.}
	\label{tab:hw_parameters}
	\vspace{-10pt}
	\centering
	{\fontsize{6}{10}\selectfont
		\begin{tabular}{lp{5cm}}
			\hline
			Neuron technology & CMOS or FinFET\\
			Technology node & 65nm, 45nm, 32nm, and 16nm\\
			Supply voltage & 1.0V\\
			Energy per spike & 50pJ at 30Hz spike frequency\\
			\hline
			Synapse technology & OxRRAAM or PCM\\
			Access device & Diode or FET or NMOS\\
			Resistance states & 1-bit/synapse or 2-bit/synapse\\
			\hline
	\end{tabular}}
	\vspace{-10pt}
\end{table}

The generalized template of Tables~\ref{tab:parasitics} and \ref{tab:hw_parameters} can be infused with the specific details of a present-day neuromorphic chip and evaluate the impact of technology scaling on system-level metrics such as energy, latency, and reliability. We now present the evaluation of \tech{} by configuring it with the parameters of the DYNAPs neuromorphic hardware~\cite{dynapse} at 45nm node with 2-bit synapses implemented using OxRRAM-based 1T1R cells.

%% file: sections/evaluation.tex
Recently, SNNs are used to improve the quality of machine learning applications~\cite{HeartEstmNN,moyer2020machine,smith2020solving,hamilton2020modeling}.
We select 10 machine learning programs which are representative of three most commonly-used neural network classes: convolutional neural network (CNN), multi-layer perceptron (MLP), and recurrent neural network (RNN).
Table~\ref{tab:apps} summarizes the topology, the number of neurons and synapses, the number of spikes per image, and the baseline accuracy of these applications on the DYNAPs neuromorphic hardware.

\begin{table}[h!]
	\renewcommand{\arraystretch}{0.8}
	\setlength{\tabcolsep}{2pt}
	\caption{Applications used to evaluate \tech{}.}
	\label{tab:apps}
	\vspace{-10pt}
	\centering
	\begin{threeparttable}
	{\fontsize{6}{10}\selectfont
		\begin{tabular}{ccc|ccl|c}
			\hline
			\textbf{Class} & \textbf{Applications} &
			\textbf{Dataset} &
			\textbf{Neurons} & \textbf{Synapses} & \textbf{Avg. Spikes/Frame} & \textbf{Accuracy}\\
			\hline
			\multirow{6}{*}{CNN} & LeNet & MNIST & 23,687 & 275,110 & 724,565 & 85.1\%\\
			& DenseNet & CIFAR-10 & 17,450 & 198,470 & 1,250,976 & 42.8\%\\
			& AlexNet & ImageNet & 259,604 & 3,873,222 & 7,055,109 & 69.8\%\\
			& ResNet & CIFAR-10 & 267,488 & 35,391,616 & 7,339,322 & 57.4\%\\
			& VGG & ImageNet & 623,635 & 12,215,209 & 12,826,673 & 90.7 \%\\
			& HeartClass~\cite{das2018heartbeat} & Physionet & 170,292 &  1,049,249 & 2,771,634 & 63.7\%\\
			\hline
			\multirow{3}{*}{MLP} & MLPDigit & MNIST & 894 & 79,400 & 26,563 & 91.6\%\\
			& EdgeDet \cite{carlsim} & CARLsim & 7,268 &  114,057 & 248,603 & 100\%\\
			& ImgSmooth \cite{carlsim} & CARLsim & 5,120 & 9,025 & 174,872 & 100\%\\
			\hline
 			RNN & RNNDigit \cite{Diehl2015} & MNIST & 1,191 & 11,442 & 30,508 & 83.6\%\\
			\hline
	\end{tabular}}
	\end{threeparttable}
\end{table}

We evaluate the following three configurations of \tech{}.
\begin{itemize}
    \item \textbf{\pc{}}~\cite{pycarl}: This is our default configuration, where a machine learning model is clustered arbitrarily. Clusters are also mapped arbitrarily to the crossbars of a hardware.
    \item \textbf{\sm{}}~\cite{spinemap}: In this configuration, \tech{} clusters a machine learning model to minimize the inter-cluster spike communication. Clusters are mapped to the crossbars to reduce energy consumption on the shared interconnect.
    \item \textbf{\esl{}}~\cite{esl20}: In this configuration, \tech{} decomposes a machine learning model to pack its neurons and synapses densely into each crossbar. The clusters are mapped to the crossbars to reduce spike latency and energy consumption on the shared interconnect.
\end{itemize}

\subsection{Software Exploration: Cluster Count}
\begin{figure}[h!]
	\centering
	\vspace{-10pt}
	\centerline{\includegraphics[width=0.99\columnwidth]{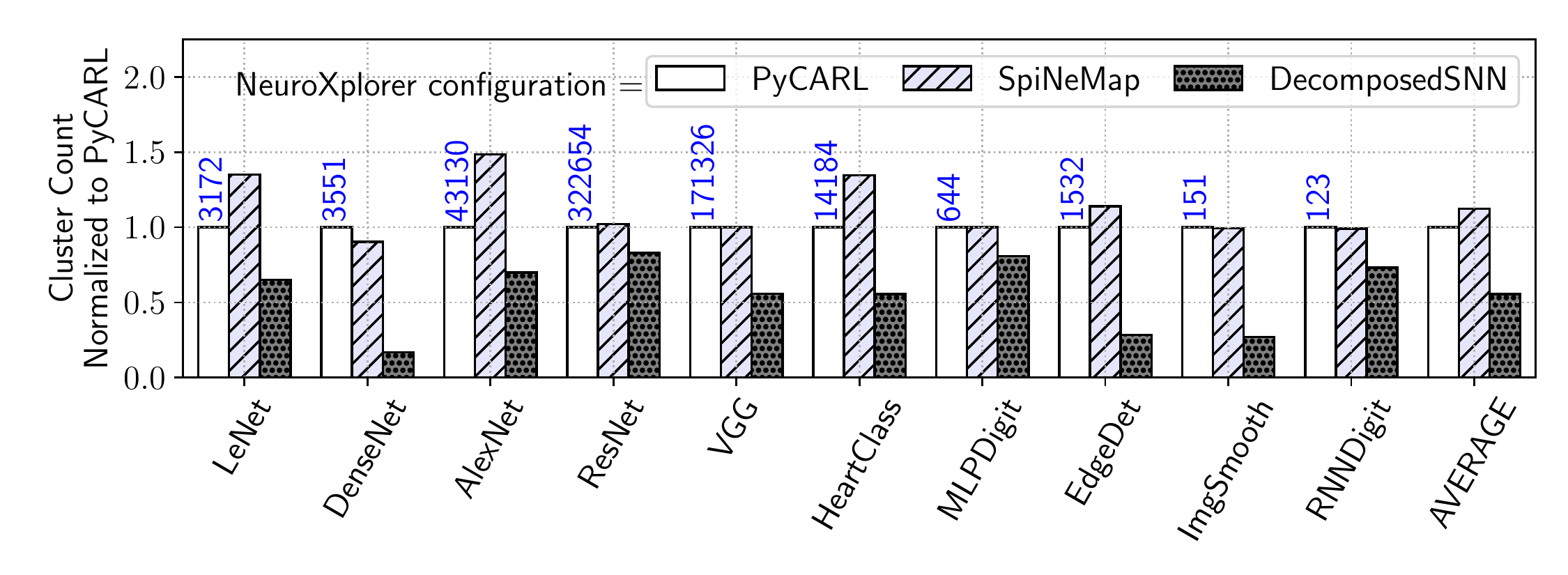}}
	\vspace{-10pt}
	\caption{Cluster count using \tech{}.}
	\vspace{-10pt}
	\label{fig:cluster_count}
\end{figure}

Figure~\ref{fig:cluster_count} plots the cluster count for each evaluated application for three different configurations of \tech{}, normalized to \pc{}. For reference, the number of clusters obtained using \pc{} is indicated for each application. We observe that different configurations of \tech{} lead to different cluster counts. \esl{}, which maximizes the neuron and synapse utilization within each cluster, generates the lowest cluster count (44.5\% lower than \pc{} and 50.1\% lower than \sm{}). 

\subsection{Software Exploration: Spike Count}
\begin{figure}[h!]
	\centering
	\vspace{-10pt}
	\centerline{\includegraphics[width=0.99\columnwidth]{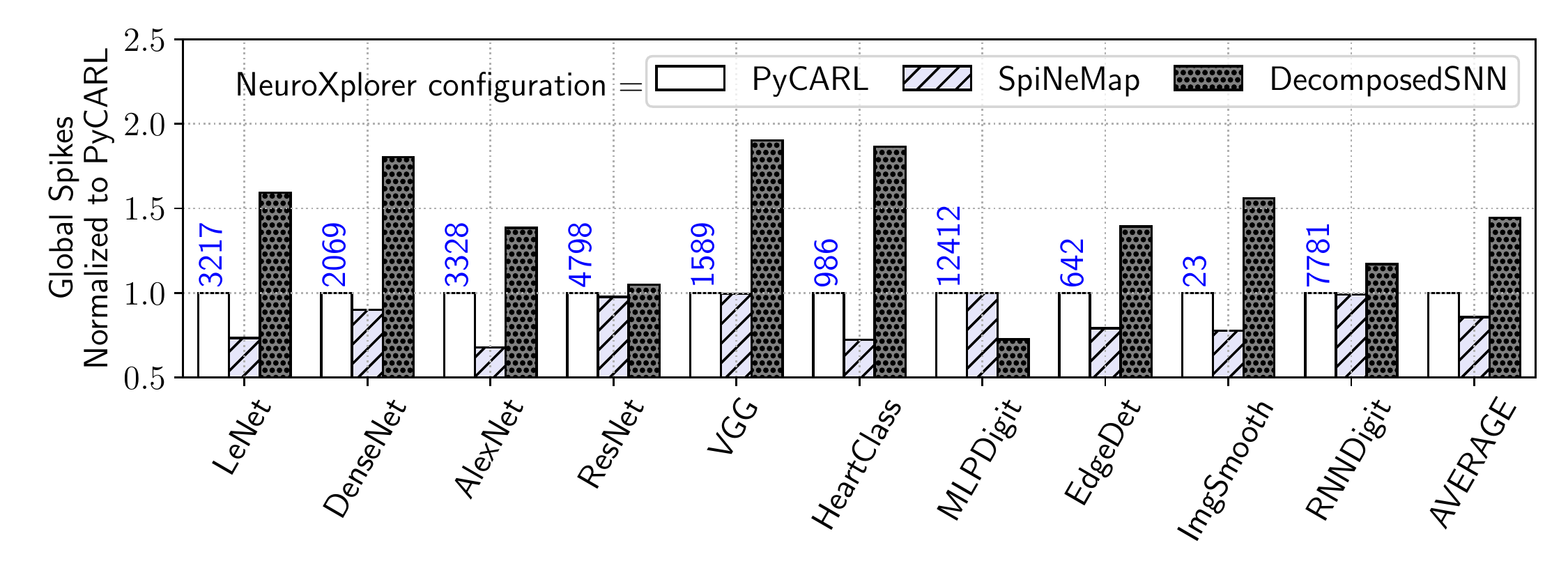}}
	\vspace{-10pt}
	\caption{Global spikes using \tech{}.}
	\vspace{-10pt}
	\label{fig:global_spikes}
\end{figure}

Figure~\ref{fig:global_spikes} plots the total number of spikes on the shared interconnect (called global spikes) for each evaluated application for three different configurations of \tech{}, normalized to \pc{}. We observe that \sm{} has the lowest number of global spikes (6\% lower than \pc{} and 34\% lower than \esl{}), which reduces both spike latency and communication energy due to reduction of the congestion on the interconnect.

\subsection{Hardware Exploration: Energy and ISI}
\begin{figure}[h!]
	\centering
	\centerline{\includegraphics[width=0.99\columnwidth]{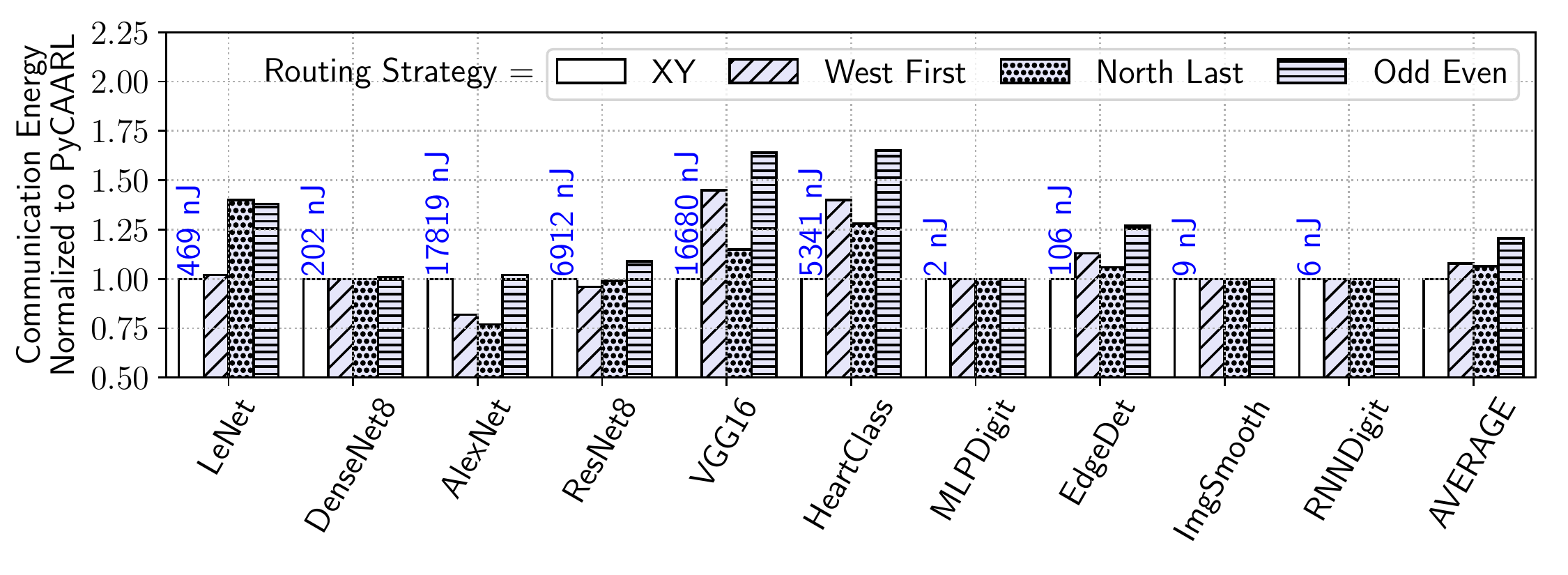}}
	\vspace{-10pt}
	\caption{Communication energy using \tech{}.}
	\label{fig:energy}
\end{figure}

\begin{figure}[h!]
	\centering
	\vspace{-10pt}
	\centerline{\includegraphics[width=0.99\columnwidth]{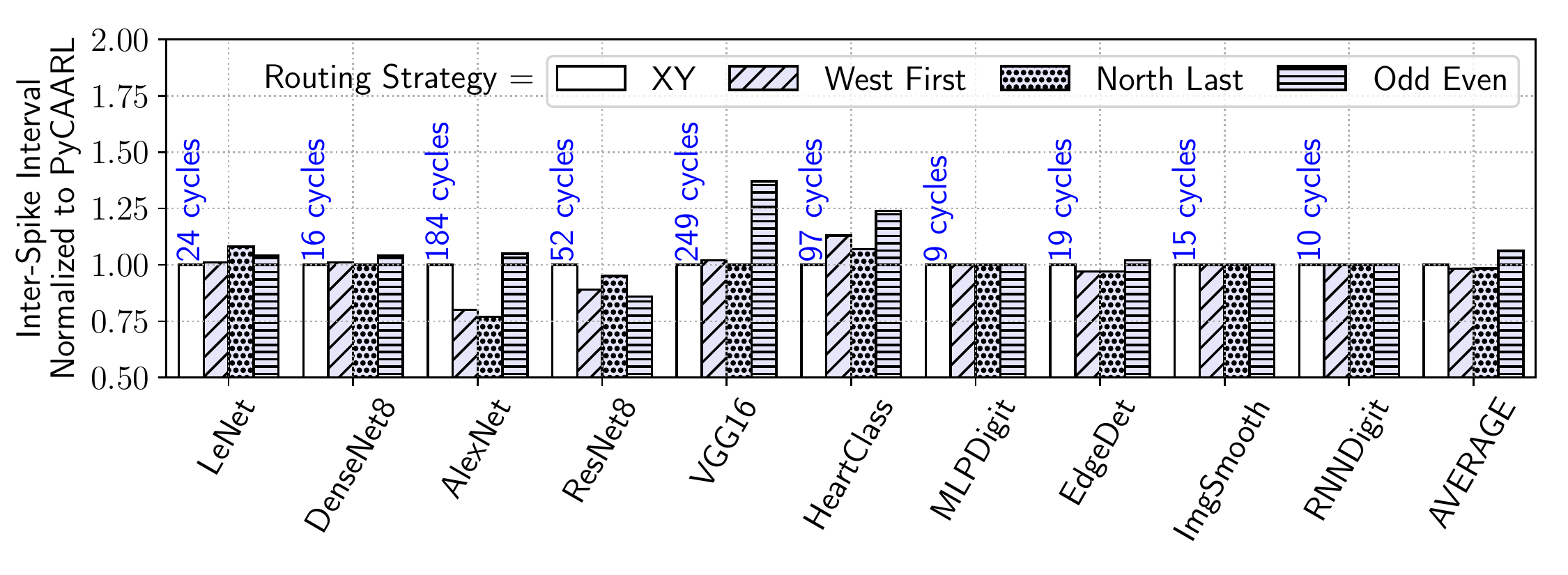}}
	\vspace{-10pt}
	\caption{Inter-spike interval (ISI) using \tech{}.}
	\label{fig:isi}
\end{figure}

Figure~\ref{fig:energy} and \ref{fig:isi}  plot respectively, the communication energy and inter-spike interval (ISI) of each evaluated application using four NoC routing techniques of \tech{} normalized to XY routing. For reference, the communication energy and ISI at 45nm technology node with XY routing is also indicated.

\subsection{Technology Exploration: Inference Lifetime}
\begin{figure}[h!]
	\centering
	\vspace{-10pt}
	\centerline{\includegraphics[width=0.99\columnwidth]{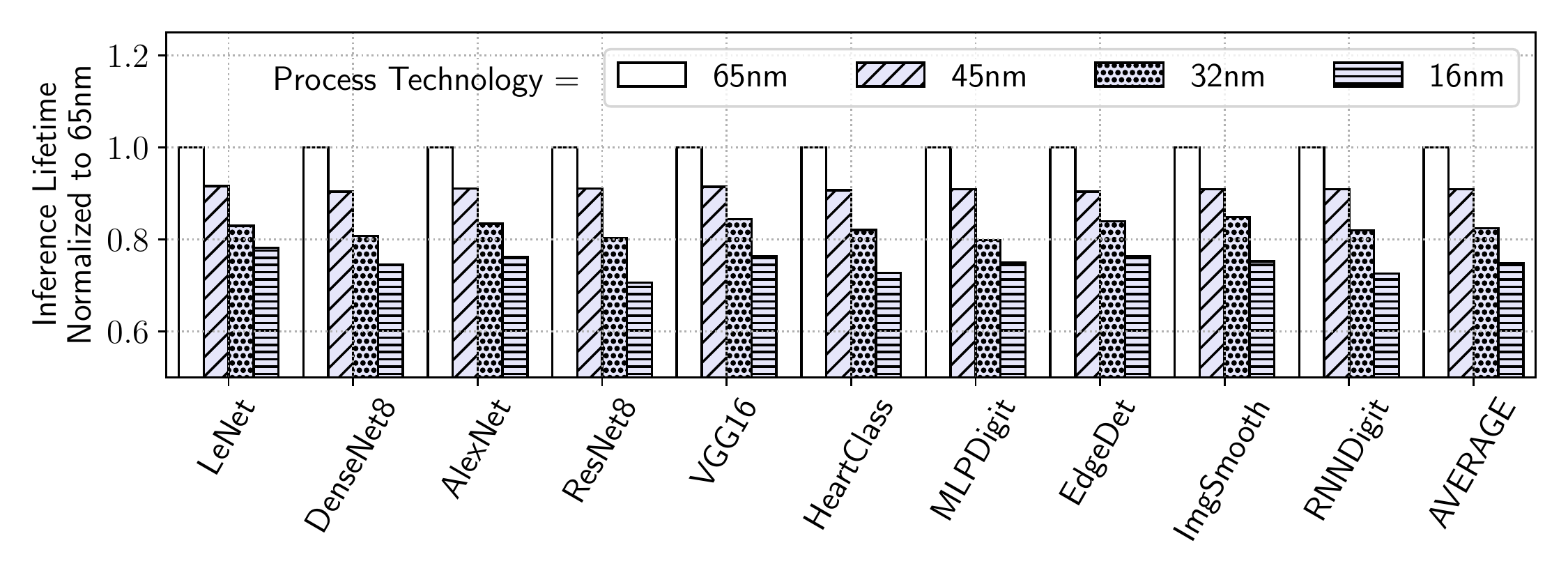}}
	\vspace{-10pt}
	\caption{Technology exploration using \tech{}.}
	\vspace{-10pt}
	\label{fig:technology}
\end{figure}

Non-Volatile Memories (NVMs) suffer from read endurance problem, where an NVM cell can switch its state upon repeated access. Therefore, the programmed synaptic weights of the NVM cells in a neuromorphic hardware needs to reprogrammed periodically. To this end, inference lifetime refers to how many images can be successfully inferred using the hardware before reprogramming of the synaptic weights becomes necessary. Figure~\ref{fig:technology} plots the impact of technology scaling on the inference lifetime of a neuromorphic hardware. At scaled nodes, the read endurance of NVMs reduces, which lowers the inference lifetime.

%% file: sections/conclusions.tex
We propose \tech{}, an extensible framework for architectural exploration with Spiking Neural Networks (SNN). \tech{} is based on a generalized template and can be infused with specific details of a neuromorphic hardware and technology. \tech{} can perform platform-based design, hardware-software co-design, and design-technology co-optimization,
enabling system designers to explore a variety of both application as well as platform design configurations to meet the needs of emerging workloads as well as newer design technologies.
In addition to these architecture-centric functionalities, \tech{} also facilitates functional simulations via SNN simulators 
supporting different degrees of neuro-biological details and learning modalities, allowing exploration of application quality.